\documentclass[review]{elsarticle}

\usepackage{lineno,hyperref}
\usepackage[english]{babel}
\usepackage[utf8x]{inputenc}
\usepackage{amsmath,amssymb}
\usepackage{bm}
\usepackage{graphicx}
\graphicspath{{Figures/}}
\usepackage{siunitx}
\usepackage{booktabs}
\usepackage{mathtools}

\usepackage{caption}
\usepackage{numprint}
\usepackage{multirow}
\usepackage[ruled,linesnumbered]{algorithm2e}
\usepackage{chngcntr}
\usepackage{parskip}
\usepackage{mathtools}

\modulolinenumbers[5]
\usepackage[top=30mm, bottom=25mm, right=30mm, left=30mm]{geometry}
\usepackage{xcolor}

\begin{document}

\begin{frontmatter}

\title{Grey-box Models for wave loading prediction}

\author[Address1]{D.J. Pitchforth\corref{mycorrespondingauthor}}
\cortext[mycorrespondingauthor]{Corresponding author}
\ead{djpitchforth1@sheffield.ac.uk}

\author[Address1]{T.J. Rogers}
\author[Address2]{U.T.Tygesen}
\author[Address1]{E.J. Cross}

\address[Address1]{Dynamics Research Group, Department of Mechanical Engineering, University of Sheffield, Mappin Street, Sheffield S1 3JD, United Kingdom}
\address[Address2]{Ramboll Energy, Global Division-Jackets, Bavnehøjvej 5, DK-6700 Esbjerg, Denmark}

\begin{abstract}
The quantification of wave loading on offshore structures and components is a crucial element in the assessment of their useful remaining life. In many applications the well-known Morison’s equation is employed to estimate the forcing from waves with assumed particle velocities and accelerations. This paper develops a grey-box modelling approach to improve the predictions of the force on structural members. A grey-box model intends to exploit the enhanced predictive capabilities of data-based modelling whilst retaining physical insight into the behaviour of the system; in the context of the work carried out here, this can be considered as physics-informed machine learning. There are a number of possible approaches to establish a grey-box model. This paper demonstrates two means of combining physics (white box) and data-based (black box) components; one where the model is a simple summation of the two components, the second where the white-box prediction is fed into the black box as an additional input. 
Here Morison’s equation is used as the physics-based component in combination with a data-based Gaussian process NARX - a dynamic variant of the more well-known Gaussian process regression. Two key challenges with employing the GP-NARX formulation that are addressed here are the selection of appropriate lag terms and the proper treatment of uncertainty propagation within the dynamic GP. The best performing grey-box model, the residual modelling GP-NARX, was able to achieve a 29.13\% and 5.48\% relative reduction in NMSE over Morison's Equation and a black-box GP-NARX respectively, alongside significant benefits in extrapolative capabilities of the model, in circumstances of low dataset coverage.
\end{abstract}


\end{frontmatter}


\section{Introduction}
Many engineering structures within the North Sea have already exceeded or are close to their initially specified 20-25 year design lives \cite{NorthSeaStruct}. The financial incentive for continued operation of structures, beyond their design lives, brings attention to the need for accurate prediction of remaining fatigue life. Safety concerns around the operation of ageing structures are a key priority and confidence within prognosis is, therefore, paramount.

Offshore engineering structures operate in harsh environments, in which accurate representation of dynamic behaviour and prediction of remaining fatigue life is difficult. In such extreme environments, physics-based (white-box) models are often unable to fully capture the complexity of dynamic behaviour. For example, phenomena including movement of mechanical joints, thermal effects and humidity are difficult to characterise and, therefore, model and validate in a dynamic context.

A data-based (black-box) approach aims to provide a more flexible alternative, where machine learning techniques may be used to characterise relationships between variables directly from data. The nature of the variables being modelled is arbitrary and no prior understanding of the physics is required. Although purely data-based approaches have proven to be effective in the prediction of structural responses in changing environments \cite{GPStrainLizzy,ARARXSHM,AspectsSHMCMTurbines}, machine learning models have their drawbacks. Black-box models are generally poor at extrapolation, with performance suffering in conditions outside the scope of provided training data. Overfitting and the adoption of unnecessarily complex model structures during training can also be an issue \cite{BishopPatternRecML,HawkinsOverfit}. 
 
A grey-box approach combines physics-based modelling and data-based learning with the aim of having a flexible model that is informed by physical insight. A key area of expected improvement concerns extrapolation; it is hoped that the structured white-box component of the model will assist inference in areas where training data coverage is low. Improvement of performance outside the observed training conditions would reduce the current demand for increasingly large training datasets and associated monitoring efforts.

The dataset used within this paper was collected from the Christchurch Bay Tower (CBT), an offshore test facility constructed with the intention to develop better understanding of wave and current action \cite{ChristchurchProject}. The tower was equipped with Perforated ball Velocity Meters (PVMs) to measure water particle velocity and acceleration, along with force sleeves to measure the horizontal force acting on the tower. This provides a dataset capturing a real sea state environment with valuable measurements of the modelling target, wave loading force, which allows for the validation and performance measurement of modelling approaches.

Within research communities, the study of Computational Fluid Dynamics (CFD) has dominated the quantification of wave loading forces \cite{CFDWaveLoadCoastalBridge,CFDWaveImpact}. Within industrial applications however, the high computational resource requirements of CFD and difficulty of model validation for structures in complex environments, has led to a preference, in some industries at least, for more simplistic empirical methods. A common example of one such method is Morison's Equation \cite{Morison1950}, which offers an empirical solution for wave loading with minimal computational resources. This paper will explore methods of combining Morison's Equation with black-box Gaussian process and Gaussian process NARX models. The aim being improving predictive performance, still within a reasonable computational budget.

\section{Model Architectures}
This section details the modelling methodology proposed for the prediction of wave loading. Morison's Equation forms the basis for the white-box model construction (with Bayesian linear regression for parameter estimation), whilst a GP-NARX model is used for the black-box. Methods of combining the approaches to form grey-box models are also presented in this section.

\subsection{White-box}
Morison's equation has been a widely used tool for the modelling of wave loading on slender members since its introduction in 1950 \cite{Morison1950}, being used in applications including wind turbine design \cite{MorisonAsumpSlender} and characterising dynamic behaviour of offshore spar platforms \cite{MorisonSpar}. Its popularity has been helped, in part, by benefits in computation time over more complex, CFD approaches. To achieve such benefits, Morison's Equation relies on a number of simplifying assumptions:
\begin{itemize}
  \item The waves are not affected by the presence of the submerged members. For a cylindrical structure, the wavelength and water depth should far exceed the diameter \cite{MorisonAsumpSlender}.
  \item Flow should be unidirectional \cite{MorisonUnidirectional}.
  \item The wave force may be separated in to a velocity-dependant drag term and an acceleration-dependant inertia term, simplifying the wave-structure interaction \cite{MorisonUnidirectional}.
  \item The considered waves are surface waves and unbroken \cite{Morison1950}.
\end{itemize} 
The simplest form of Morison's equation assumes the structure on which the wave force acts is a rigid, slender cylinder. For a given wave velocity \(U\) and acceleration \(\dot{U}\), the force per unit axial length \(F\) is given as:
\begin{equation}
F = \underbrace{\frac{1}{2}\rho DC_d}_{\begin{matrix}{C_d'}\end{matrix}}U|U| + \underbrace{\frac{1}{4}\pi\rho D^2C_m}_{\begin{matrix}{C_m'}\end{matrix}}\dot{U}
\end{equation}
where \(\rho\) is the fluid density, \(D\) is the cylinder diameter, \(C_d\) is the drag coefficient and \(C_m\) is the inertia coefficient.
The dimension specific terms may be grouped to form two constants \(C_d'\) and \(C_m'\) relating to the drag and inertia forces of the wave. This leads to the simplified version of Morison's equation used within this paper:
\begin{equation}
F = C_d'U|U| + C_m'\dot{U}
\end{equation}
The construction of the white-box model presented will rely on the estimation of the grouped parameters within Morison's equation. The simplicity of this approach will aid in minimising the complexity of the final combined grey-box models presented in later sections, although a non-dimensional form can be readily used.

Parameter estimation and model prediction with Morison's equation is achieved via Bayesian linear regression, an introduction to which is provided in Appendix A.1. Approaching the regression in a Bayesian manner provides distributions over the parameter estimates and confidence intervals for the predictions, which can then be compared with the Gaussian process models used later. 

\subsection{Black-box}

Gaussian Process Regression (GPR), utilised here, is a non-parametric, flexible, Bayesian machine learning technique \cite{GPRasmussen}. The return of confidence intervals with predictions, minimal requirement for prior knowledge and modelling capabilities under the presence of noise have lead to the popularity of GPR within a wide range of usage applications. These span from standard regression tasks \cite{GPsForBigData}, to image processing \cite{GPImageProcessing}, to more engineering relevant examples \cite{GPMagnetic,GPStrainLizzy,rogers2020probabilistic}. A dynamic-variant of a GP regression model is employed here, namely a GP-NARX \cite{NARXWaveForce,ConfidenceBoundsNARX,DynamicSystemsGPNARX}.

A Nonlinear AutoRegressive model with eXogenous inputs (NARX) is a function of previous signal values and additional (exogenous) inputs, in which both are fed through some nonlinear function \(f(x)\).
\begin{equation}
y_t = f([u_{t},\ u_{t-1}, ...,\ u_{t-l_u},\ y_{t-1},\ y_{t-2}, ...,\ y_{t-l_y}]) + \varepsilon
\end{equation}
The previous signal values, \(y_{t-i}\) and exogenous inputs, \(u_{t-j}\) are considered up to \(l_{y}\) and \(l_{u}\) lagged time steps respectively. For the wave force estimation in this paper, the exogenous inputs, \(u\), considered are the velocity, \(U\) and acceleration, \(\dot{U}\) of the wave particles.

The nonlinear function, $f(x)$, in a NARX model is commonly fixed to be a polynomial \cite{PolyNARXWienerHammerstein,PolyNARXSimErr}, but in the case of a GP-NARX, a Gaussian process (GP) is used. With a GP one avoids needing to fix the functional form explicitly, instead, the selection of a mean and covariance function defines a family of feasible functions that may explain the data. An overview of basic GP theory is given in Appendix A.2, with the reader encouraged to consult \cite{GPRasmussen} for a more detailed understanding.

\subsubsection{One step ahead and model predicted output}
There are two types of prediction that may be calculated from any autoregressive model form: One Step Ahead (OSA) and Model Predicted Output (MPO)\footnote{In some communities these are referred to as the \emph{prediction} and \emph{simulation} tasks for OSA and MPO respectively.}. For OSA, previously measured values of the output signal are used as lagged inputs to the model, whilst MPO requires the feedback of the model prediction itself. The MPO performance will generally be worse than that of OSA due to the compounding of model errors, however, it is a much more representative measure of how well the model has captured the true dynamics of the process and therefore a more rigorous test.

The practical use of an OSA prediction occurs most naturally in a control setting, where continual measurements of the target of interest are available. In a Structural Health Monitoring (SHM) context, the assumption is that continual measurement of the wave force itself will not be available, meaning that an OSA prediction will not generally be useful. This paper, therefore, focuses on the MPO task.

The focus on an MPO necessitates careful attention to how the hyperparameters in the GP covariance function are optimised. In a standard static implementation of a GP, the hyperparameters, $\bm{\theta}$ of the covariance function, which control things like the roughness of predictions, are set by optimising a negative log marginal likelihood:
\begin{equation}
 \bm{\hat{\theta}} = \underset{\bm{\theta}}{\arg\max} \left\{ -log\,p(\bm{y}|X,\bm{\theta}) \right\}
\end{equation}
where $\bm y$ are the set of targets in the training set, with corresponding inputs $X$. This optimisation doesn't reflect the dynamic nature of the GP-NARX and strongly favours the performance of OSA predictions if used. 

The cost function should always be aligned with the desired performance criteria of the model, in this case the MPO. Here, therefore, a more appropriate choice of cost function is the Negative Log Predictive Likelihood of the Model Predicted Output (MPO NLPL):
\begin{equation}
\bm{\hat{\theta}} = \underset{\bm{\theta}}{\arg\max} \left\{ -log\,p(\bm{y}|\mathbb{E}(\bm{y^*}),\mathbb{V}(\bm{y^*}),\bm{\theta}) \right\}
\end{equation}
The NLPL of the MPO is calculated as a joint Gaussian likelihood of each measured data point \(y_t\) coming from the corresponding predictive distribution \(y^*_t\sim \mathcal{N}(\mathbb{E}(y^*_t),\mathbb{V}(y^*_t))\) of the GP-NARX output (the full formulation of this is shown in Appendix A.2 for the interested reader). The authors suggesting using an independent validation dataset for this step. The process is defined in Algorithm \ref{Alg:MPONLPL}. 

\IncMargin{1em}
\begin{algorithm}[ht]
\SetAlgoLined
Calculate training set covariance matrix \(K(X,X)\) for hyperparameters \(\bm{\theta}\)\\
Initialise GP-NARX for validation set from \(\bm{U_{t:t-l_u}}\) , \(\bm{\dot{U}_{t:t-l_u}}\) and \(\bm{y_{t-1:t-l_y}}\)\\
\For{\(t = 1 : T\)}
{Calculate: \(p(y^*_t | \bm{U_{t:t-l_u}}, \bm{\dot{U}_{t:t-l_u}}, \mathbb{E}(\bm{y^*_{t-1:t-l_y}}), \bm{\theta}) = \mathcal{N}( \mathbb{E}(y^*_t),\mathbb{V}(y^*_t))\)}
\(NLPL = -\sum_{t=1}^T \log \mathcal{N}(y_{t} \vert \mathbb{E}(y^*_t),\mathbb{V}(y^*_t))\)
\caption{MPO NLPL cost function for GP-NARX optimisation.}
\label{Alg:MPONLPL}
\end{algorithm}
\DecMargin{1em}

where \(\bm{U_{t:t-l_u}}\) and \(\bm{\dot{U}_{t:t-l_u}}\) refer to the lagged exogenous input vectors of velocity and acceleration, \(\bm{y_{t-1:t-l_y}}\) refers to the lagged vector of measured wave force and \(\bm{y^*_{t-1:t-l_y}}\) the lagged vector of predicted wave force.

\subsubsection{Uncertainty propagation in a GP-NARX}
One of the benefits of using a Bayesian regression approach, such as GPR, is access to the full posterior distribution and therefore availability of confidence intervals on any prediction made. This causes an issue, however, in a NARX MPO setting, as model predictions get fed back and used as model inputs at the next step. Previous uses of GP-NARX models have generally avoided the tricky issue of uncertainty propagation in an MPO setting.

Without alteration, the confidence intervals of the GP-NARX fail to account for the full uncertainty within the prediction. The MPO requires the feedback of the model prediction, for use as subsequent lagged output. This is typically taken as a point estimate of the expected value of the GP-NARX prediction, failing to acknowledge that the output of the model is in fact a distribution. This does not account for the potential variation in the feedback of model outputs which would have a cumulative effect over time. This causes an uncertain input problem in the GP which is hard to compute.

\begin{figure}[ht]
  \centering
      \includegraphics[width=1\textwidth]{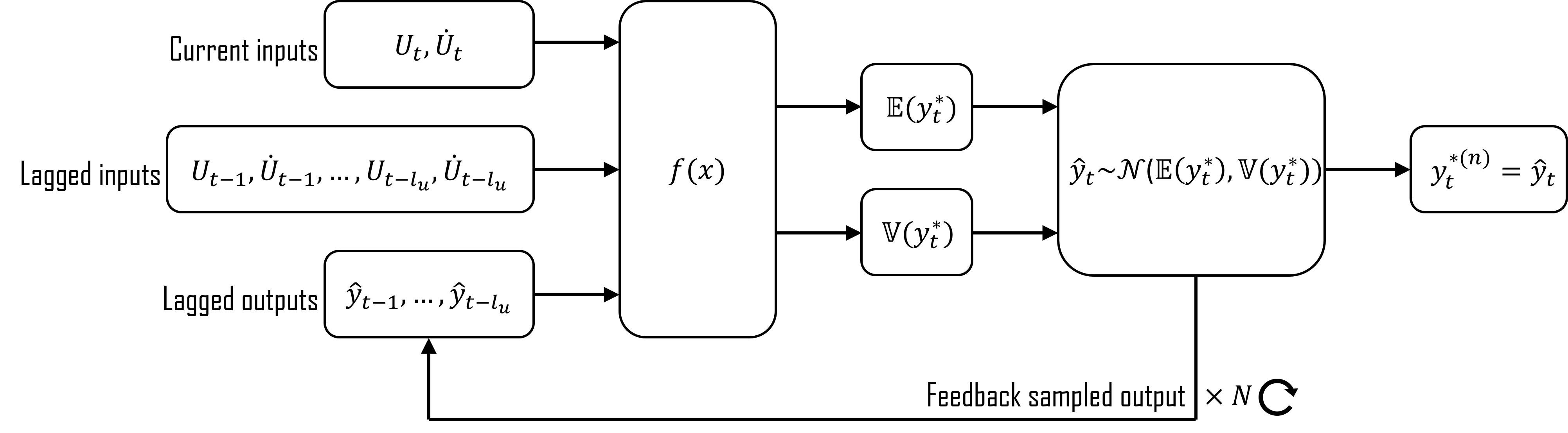}
  \caption{Block diagram of Monte Carlo uncertainty propagation within the GP-NARX.}
  \label{fig:MC_NARX_Block_Diagram}
\end{figure}

Uncertainty propagation within the GP-NARX can be achieved via the use of Monte Carlo sampling \cite{ConfidenceBoundsNARX,TimThesis}. Instead of feeding back the model output mean, a sample from the output distribution \(\hat{y}_t\) is used. This may be repeated for \(N\) samples to form a series of \(N\) potential realisations for the model output \(y_{t*}^{(n)}\) from which more realistic posterior distributions may be estimated. The procedure for generating the Monte Carlo sampled Model Predicted Output (MC MPO) is summarised in the block diagram within Figure \ref{fig:MC_NARX_Block_Diagram}.

\subsection{Grey-box}
A grey-box model combines physics and data-based approaches with the aim of extracting benefits from each of the model types: structure, insight and extrapolative performance from the white-box component and flexibility and ability to model unknown phenomena from the black-box component. There are two potential architectures presented here, both of which combine the earlier discussed Morison's equation with GP and GP-NARX models.
\subsubsection{Residual modelling}
Perhaps the simplest approach to grey-box modelling is to sum the predictions of a white and black-box model. If the white-box takes a fixed form, this summation is equivalent to using the black-box to model the residual error between the white-box and any collected data. In the FE modelling community the practice of using a machine learner (often a GP) to model the residuals from an FE model is often referred to as `bias correction', acknowledging that there is likely to be some error in the complex FE representation of the structure. It is possible to apply the approach to more simplistic base models such as Morison's Equation, with the modelling of polynomial regression residuals using a GP being explored as early as 1975 \cite{BlightOttResidual}.

It is known that Morison's Equation simplifies the behaviour of wave loading, not accounting for effects such as vortex shedding or other complex behaviours \cite{MorisonVortex} and will typically have residual errors in the region of \(20\%\)\cite{CoastalHydraulics}. Through the use of a black-box component, these residual errors may be modelled and the result added to the white-box, producing a model of the form: 

\begin{equation}
y_t = \!\! \underbrace{F_{mor}}_{White-box} \!\!\! + \;\: \underbrace{f([U,\ \dot{U}]) + \varepsilon}_{Black-box}
\end{equation}
Here, the parameters in Morison's equation will be established via Bayesian linear regression. The GP or GP-NARX model is set up identically to those discussed within the black-box section, except that the target is now the residual error of Morison's Equation rather than the measured wave force itself. The intention is to capture the missing physics excluded by the simplifications present within Morison's Equation.

The magnitude of the black-box term can be interpreted as being the extent to which the data confirms the prediction of Morison's Equation. In regions of high uncertainty, outside the observed training data, the GP will revert to its prior of zero with the overall model therefore outputting Morison's Equation. An equivalent view of this is the usage of a white-box mean function within the GP or GP-NARX \cite{GPRasmussen}.

\subsubsection{Input augmentation}
An alternative means by which physics and data-based approaches may be combined is via the inclusion of the white-box model output as an additional input to the black-box. Models involving the manipulation of the non-parametric components of black-boxes using physics have been termed `type B' \cite{GreyNonLinearBenchmarks} grey-box models within the nonlinear system identification community, whilst the transformation of model input data through physical insight has been referred to as `semi-physical' modelling \cite{ToolsSemiPhysical}. This approach was used in \cite{RiverSemiPhysical} to model and optimise hydroelectric power generation. Input augmentation has been shown to offer performance increases over white and black-box approaches in the context of a nonlinear cascaded tanks system, particularly in the case of an extended physical model \cite{GreyFramework}.

Here, the result of Morison's Equation is used along with the originally included water particle velocity \(U\) and acceleration \(\dot{U}\) as the input to the GP or GP-NARX. The model is of the form:
\begin{equation}
y_t = \overbrace{f([\!\!\!\!\underbrace{F_{mor},}_{White-box}\!\! U,\ \dot{U}]) + \varepsilon}^{Black-box}
\end{equation}
The output of the white-box model is carried forward to provide an input for the GP that is strongly linked to the physics of the problem.
An advantage of this approach over residual modelling is the maintaining of the signal to noise ratio. This is particularly important in cases of high fidelity physics-based models, where the residual of the model will be small in comparison to the model noise. 

\section{Case study: Christchurch Bay Tower}
\begin{figure}[ht]
  \centering
      \includegraphics[width=0.59\textwidth]{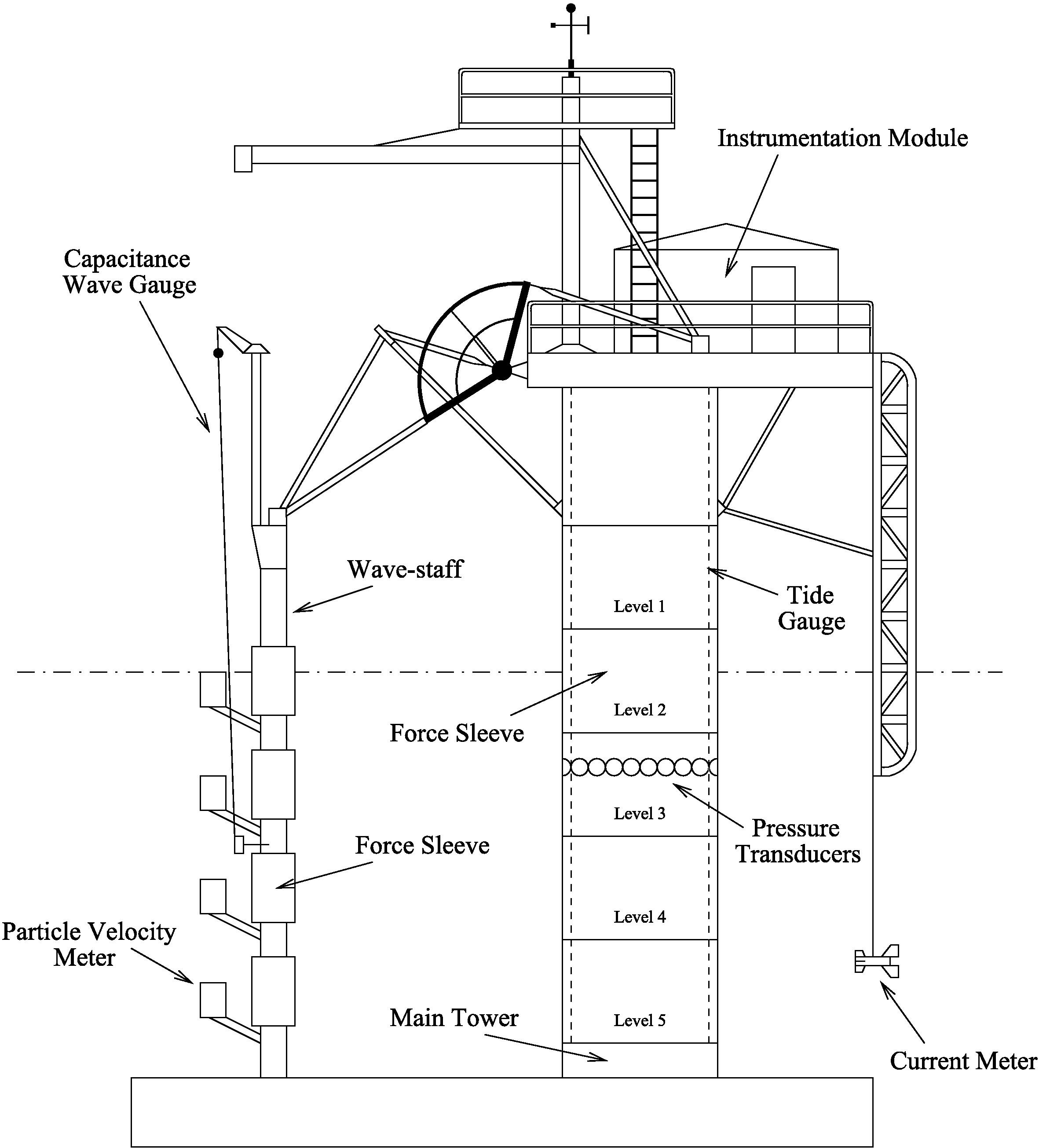}
  \caption{Schematic of Christchurch Bay Tower \cite{ConfidenceBoundsNARX}.}
  \label{fig:Christchurch_Bay_Tower}
\end{figure}

This section presents implementations of the proposed model architectures on a dataset collected from the Christchurch Bay Tower (CBT) \cite{ChristchurchProject}. This provided a test of model performance within a real sea environment. A schematic of the structure is shown in Figure \ref{fig:Christchurch_Bay_Tower}.

The structure is comprised of a large central column, 2.8m in diameter, and a smaller column, 0.48m in diameter, each equipped with an array of sensors. Perforated ball Velocity Meters (PVMs), pressure transducers, force sleeves and wave buoys were used to create a 41 channel dataset \cite{ChristchurchProject}. Although this is a historic dataset, the sensor network here is more densely populated than those that might be employed on offshore structures today due to the CBT being constructed specifically as a test facility.

Three subsets of 1000 data points, sampled at 13.25Hz, were selected from the complete dataset for use as training, validation and test sets. A region of 3000 points was selected where the ratio of the x-velocity to y-velocity of the wave was a maximum to ensure that the flow was primarily unidirectional. This was then split in to 3 sequential subsets. The study here focuses on data from the small column where the assumptions of Morison's equation around slender members are more likely to be valid.

The parameters for Morison's equation are found via Bayesian linear regression on the training dataset. The prior \(C_d\) and \(C_m\) coefficients used were obtained from Clauss \cite{Clauss1992Offshore} and DNV-RP-C205 \cite{DNVRPc205}. Note here that \(C_d\) and \(C_m\) refer to the specific drag and inertia coefficients rather than the grouped constants \(C_d'\) and \(C_m'\). The data used was taken from one of the high intensity measurement runs of the complete CBT dataset in which the Reynolds number (Re) \(>1\times10^5\) and the Keulegan–Carpenter number (Kc) was in the range \(17<Kc<26\) \cite{ChristchurchProject}. For these flow conditions, Clauss \cite{Clauss1992Offshore} suggests a drag coefficient of \(C_d=0.6\). For the inertia coefficient, DNV-RP-C205 \cite{DNVRPc205} also considers the effect of surface roughness, which for a heavily instrumented cylinder gives \(C_m=1.2\). The presence of the force sleeve, accelerometers and numerous pressure transducers along the cylinders length significantly increase the surface roughness.

\subsection{Results}
This section will present and compare the different model implementations on the CBT dataset. In all cases, the training, validation and test sets remain the same, and unless otherwise specified, all results correspond to performance on the unseen test set. The GP-NARX implementation included three lags of the autoregressive terms and one lag of the exogenous input terms; the selection of these lag terms is discussed the Appendix.

\subsubsection{GP-NARX uncertainty propagation}
\begin{figure}[ht]
  \centering
      \includegraphics[width=1\textwidth]{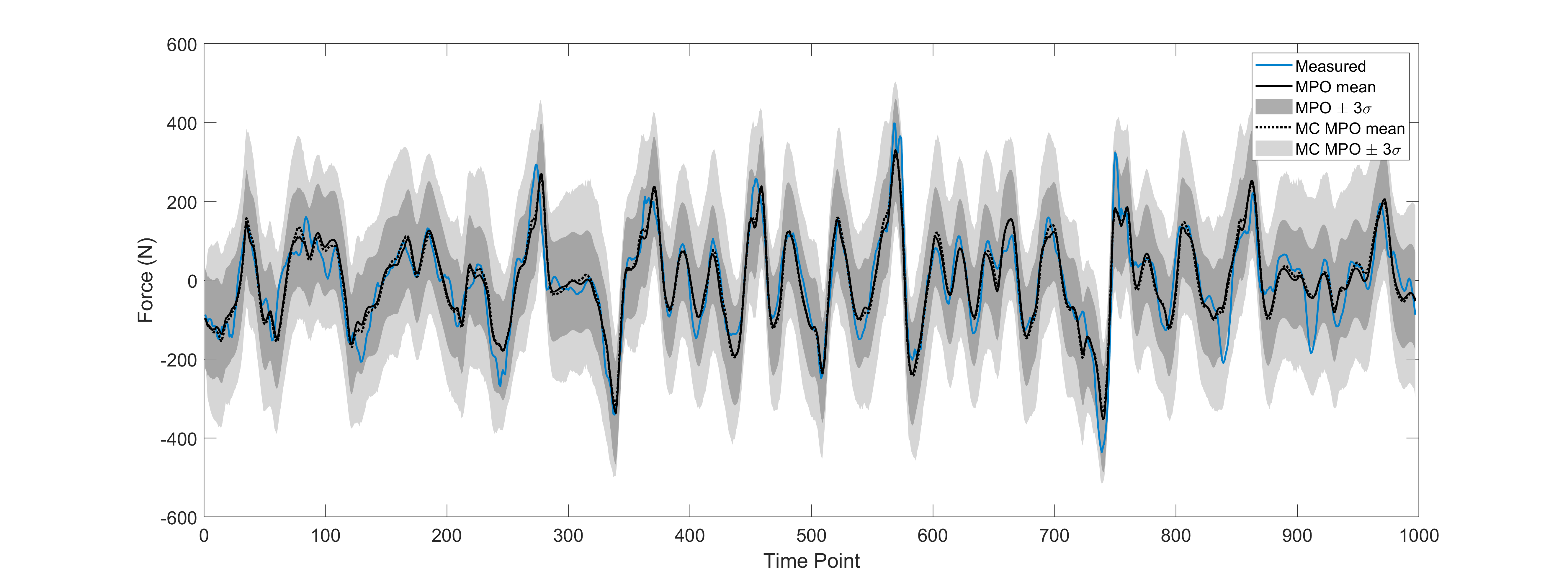}
  \caption{Comparison of black-box GP-NARX MPO and MC MPO predictions on the test set.}
  \label{fig:black_box_MPO_MC_MPO}
\end{figure}

The consideration of the uncertainty present within the feedback of GP-NARX outputs via implementation of MC MPO was found to contribute a significant amount to the overall uncertainty within predictions. Figure \ref{fig:black_box_MPO_MC_MPO} compares GP-NARX MPO and MC MPO predictions, where one can see a significant difference between the widths of confidence intervals. The average increase in \(\pm 3 \sigma\) confidence interval width was 75.0\%.

Although a perceived increase of model uncertainty may be argued a disadvantage, the underlying uncertainty present within the modelling processes has remained the same. What has instead changed is the proportion of the uncertainty that has been captured. A model should aim to be as realistic as possible about uncertainty within its predictions in order to prevent circumstances of `confidently wrong' predictions. At around time points 840 and 910, instances can be seen of poorer model performance when a sudden downwards spike in the measured data occurs. The data under study here are selected from a time when the wave state was close to unidirectional. It is likely that these spikes occur at times when the wave direction changed. This results in both a data-based component that is unable to characterise the unseen conditions well and a physics-based component where the underlying assumptions are likely less valid. In the case of the MPO prediction, the confidence bounds are not wide enough to accommodate the measured data. For the MC MPO prediction however, the confidence interval width can be seen to increase significantly in these areas and is able to account for the true behaviour. A general trend of increased uncertainty in areas of poor performance can be seen within the MC MPO.

In contrast to the confidence intervals, very little variation between the mean outputs can be seen. For the majority of the test set prediction, the MPO (solid line) and MC MPO (dotted line) are difficult to distinguish. For the purposes of uncertainty feedback, 10,000 Monte Carlo samples were used which were enough to ensure the convergence of NMSE within a $0.001\%$ tolerance and average variance within a $0.01\text{N}$ tolerance. With the wave load exceeding a $200$N magnitude in many areas, this level of precision was deemed acceptable. If the number of samples were to be increased, the mean output of MC MPO should continue to converge towards the MPO. This similarity in response means that the advantages in uncertainty representation are achieved without deterioration in performance of the mean output.

From here onwards, the MC MPO will be considered the primary prediction type of interest.

\subsubsection{Model predictive performance}
The performance of each model was assessed using the response prediction on an unseen test-set that was not used in the estimation of any parameters or hyperparameters. For the purposes of model comparison, two measures are used: the Normalised Mean Square Error (NMSE) to assess the performance of each models expected output and the Mean Standardised Log Loss (MSLL) to provide a probabilistic measure. The NMSE is expressed:
\begin{equation}
NMSE = \frac{100}{n\sigma_y^2}(\bm{y_{\star}}-\bm{y^*})^T(\bm{y_{\star}}-\bm{y^*})
\end{equation}
where \(n\) is the sample size, \(\sigma_y^2\) is the signal variance, \(\bm{y_{\star}}\) is the measured test signal and \(\bm{y^*}\) is the model prediction. An NMSE of zero implies perfect prediction whilst an NMSE of 100 would be equivalent to predicting the mean for all observations.

To construct the MSLL, one must first consider the negative log predictive likelihood of the model, \(-log\,p(\bm{y_{\star}}|X_{\star},X,\bm{y})\), where \(\bm{y_{\star}}\) is the measured test signal, \(X_{\star}\) is the set of test inputs, \(X\) is the set of training inputs and \(\bm{y}\) is the training target. Taking the negative here returns a loss rather than a utility which may be standardised by subtraction of the loss calculated when predictions equal the mean and variance of the training set. This returns a Standardised Log Loss (SLL):
\begin{equation}
SLL = -log\,p(\bm{y_{\star}}|X_{\star},X,\bm{y}) + log\,p(\bm{y_{\star}};\mathbb{E}(\bm{y}),\mathbb{V}(\bm{y}))
\end{equation}
The SLL, and therefore the MSLL, will be equal to zero for the baseline case of predicting with the training set mean and variance and increasingly negative for improved model predictions.

\begin{figure}[!ht]
  \centering
      \includegraphics[width=1\textwidth]{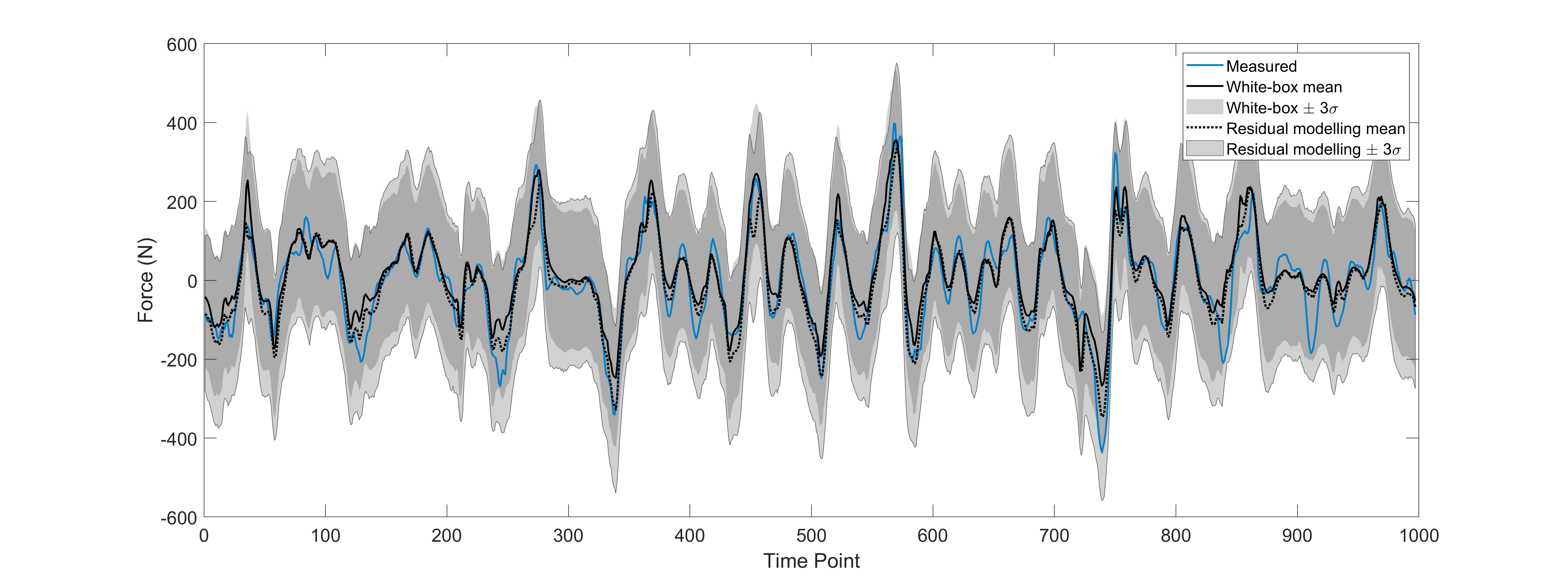}
  \caption{Test set prediction comparison between the white-box linear regression (solid line) and the MC MPO of the grey-box residual modelling GP-NARX (dotted line).}
  \label{fig:Posterior_White-box_vs_Grey-box}

    \centering
      \includegraphics[width=1\textwidth]{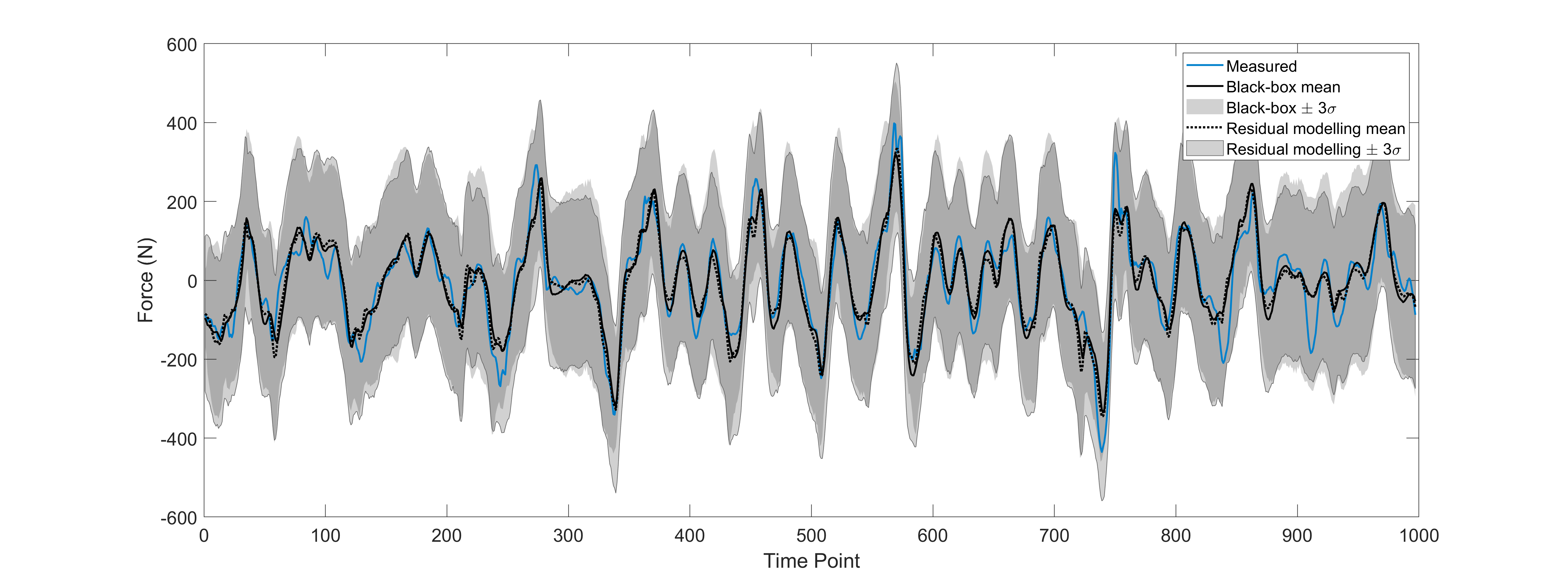}
  \caption{Test set prediction comparison between the MC MPO of the black-box (solid line) and grey-box residual modelling GP-NARX (dotted line).}
  \label{fig:Posterior_Black-box_vs_Grey-box}

  \centering
      \includegraphics[width=1\textwidth]{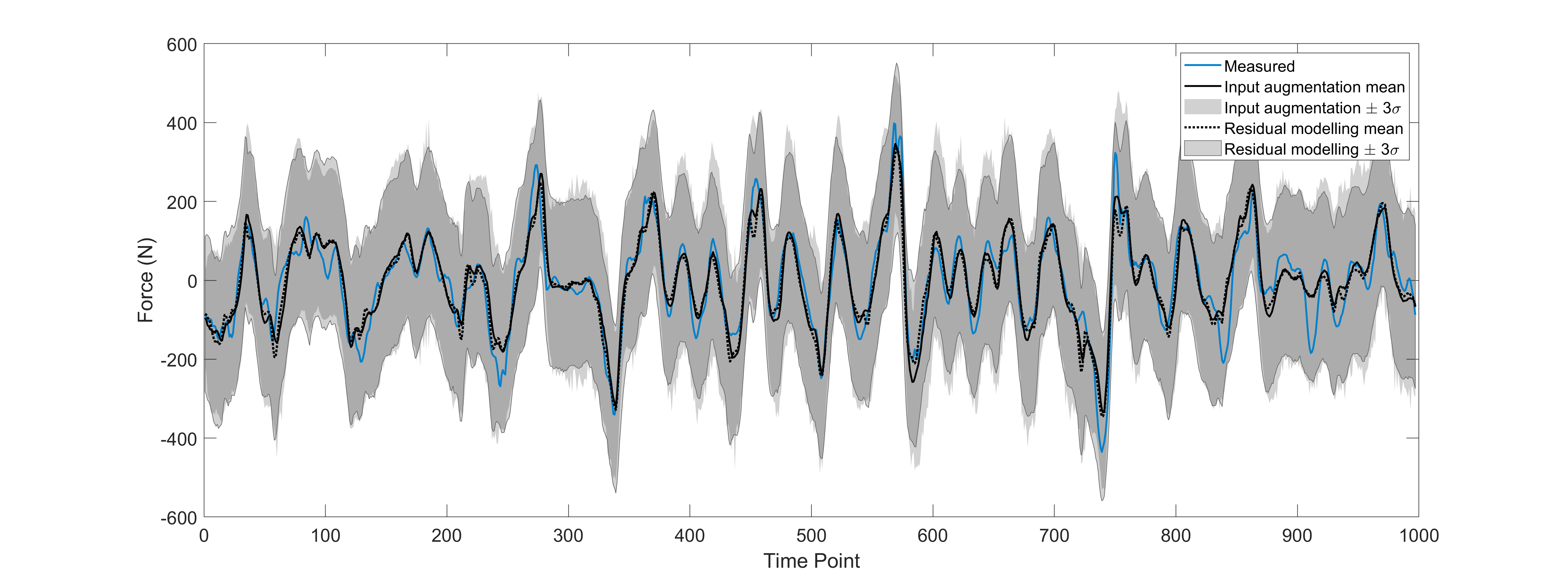}
  \caption{Test set prediction comparison between the MC MPO of the grey-box input augmentation (solid line) and residual modelling (dotted line) GP-NARX models.}
  \label{fig:Posterior_Grey-box_vs_Grey-box}
\end{figure}

A comparison of metrics for the models and their various prediction types is shown in Table \ref{tab:CompNMSE}. Comparisons of the full test set posterior between the model with the lowest NMSE, the grey-box residual modelling GP-NARX, and other model types are shown in Figures \ref{fig:Posterior_White-box_vs_Grey-box}-\ref{fig:Posterior_Grey-box_vs_Grey-box}. All results presented within this section relate to models constructed using the full training and validation sets.
\begin{table}[ht]
\sisetup{round-mode=places,round-precision=3}
  \begin{center}
    \caption{Performance comparison of model types.}
    \label{tab:CompNMSE}
    \begin{tabular}{llSS} 
    \toprule
      \textbf{Model} & \textbf{Model type} & \textbf{NMSE (\%)} & \textbf{MSLL}\\
      \midrule
      Morison's Equation  & White-box & 19.5284 & -0.8130\\
              \midrule
            \multirow{3}{*}{Static GP}& Black-box & 16.4326 & -0.9392\\
    & Residual modelling & 16.751 & -0.9142\\ & Input augmentation & 15.6272 &-0.9510\\
      \midrule
            \multirow{3}{*}{GP-NARX OSA}& Black-box & 2.7019 & -1.4441\\
    & Residual modelling & 5.212 & -1.0119\\ & Input augmentation & 2.9474 & -1.4290\\
    \midrule
                \multirow{3}{*}{GP-NARX MPO}& Black-box & 14.773 & -0.9682\\
    & Residual modelling & 13.8620 & -0.8724\\ & Input augmentation & 14.0722 & -0.9939\\
    \midrule
                \multirow{3}{*}{GP-NARX MC MPO}& Black-box & 14.643 & -0.7881\\
    & Residual modelling & 13.840 & -0.8345\\ & Input augmentation & 14.0876 & -0.7905\\
      \bottomrule
    \end{tabular}
  \end{center}
\end{table}

A stand out observation from the results in Table \ref{tab:CompNMSE} is the significant performance gap, in terms of both NMSE and MSLL, between the GP-NARX OSA and all other prediction types. This is to be expected due to the nature of OSA predictions and inclusion of lagged measured outputs within the model inputs. The prediction of a single time step ahead is of very limited use in SHM applications, particularly in the case of high sample rates, thus the good performance is of little benefit.

The NMSE of the white-box linear regression was found to be 19.528\% which is in line with the expected \(20\%\) \cite{CoastalHydraulics} residuals of Morison's Equation. Although this is around 3-6\% higher than the NMSE of other models, the result is achieved with reduced modelling complexity and computational burden. Considering the simplified version of Morison's Equation used had only the two model parameters \(C_d'\) and \(C_m'\) to model the relatively complex wave load, even moderate levels of model performance are commendable.

All models including a machine learning component were able to offer significant reductions in NMSE over the white-box model. The success of the grey and black-box models was to be expected due to the failure of Morison's Equation to account for complex behaviours present within wave loading such as vortex shedding \cite{MorisonVortex}. The inclusion of the black-box component, whether this be a GP or GP-NARX, increased model flexibility, allowing the representation of such behaviours. This indicates that previously missed underlying structure within the wave force was then able to be captured.

For all grey-box models except the residual modelling static GP, modest improvements in NMSE over the equivalent black-box approach were observed, implying that the inclusion of physics through the white-box component was able to aid model performance. For the residual modelling static GP, the inability of the black-box component to model the dynamics in the white-box residuals led to a reduction in performance. The fact that the residual modelling GP-NARX was the best performing grey-box model suggests the presence of some structure within the residuals only able to be captured by the GP-NARX and not by the static GP. In terms of missed phenomena from the model, this would be a process captured well by an autoregressive model exhibiting features such as periodicity over a small time scale.

In general, a larger increase in performance can be seen between the grey-box and white-box models, than between the grey-box and black-box models. The primary reason for this is when the full training and validation sets are used, the model is deemed to be mostly interpolating (See Section 3.1.3). Black-box and grey-box models are expected to achieve a similar performance in interpolation, with a physics-based component being most useful to assist with extrapolation. The specific type of components used to construct a grey-box model will also affect the relative differences in performance. Here, Morison’s Equation – an approximate wave loading solution, is combined with a GP-NARX – a relatively powerful black-box architecture.  The computational balance is heavily weighted in favour of the GP-NARX and it is to be expected that the performance of the grey-box would be more similar to the black-box than the white box in this case.

The MSLL of Morison's Equation was \(-0.813\), which being in line with the MSLL of the residual modelling GP-NARX MC MPO of \(-0.835\), indicated a strong model performance. Although a similar MSLL results in a similar prediction likelihood, it does not describe other aspects of model performance. Morison's Equation achieved the result with the highest NMSE of all models and the residual modelling GP-NARX the lowest. The similar MSLL was achieved through the lower prediction variance of Morison's Equation which can be seen from the narrower confidence interval width within Figure \ref{fig:Posterior_White-box_vs_Grey-box}. The trade-off between variance and NMSE would generally be preferred in favour of NMSE with wider confidence intervals better able to contain the measured result.

The earlier discussed effect of GP-NARX uncertainty propagation can be seen within the increase in MSLL between the MPO and MC MPO of the GP-NARX. The increased prediction variance caused by the feedback of output distribution samples reduced the likelihood of the prediction significantly. However, the proper treatment of uncertainty is important in preventing overestimation of prediction likelihood and overconfidence within predictions far from the observation.

\subsubsection{Model performance during extrapolation}
A major drawback of black-box models is poor performance in conditions outside those experienced within the training and validation sets. In an engineering context, this necessitates the collection of data across all possible operating conditions of the structure or system of interest. Such a demand may be extremely costly, or impossible in some cases. Improvements in extrapolative capabilities would relax the dataset coverage of conditions required for effective machine learning implementation, thereby reducing the associated monitoring efforts. This is where one would expect a grey-box model to be of particular use.

To consider the extent to which a model is extrapolating, it is useful to consider how the training, validation and test sets overlap within the input space, which for the case of wave loading is formed of the velocity \(\bm{U}\) and acceleration \(\bm{\dot{U}}\) of the wave particles, and their lags. In the idealised case, all conditions within the test set will lie within the area/volume\footnote{In most cases this is likely to be an $n-$dimensional volume, as the input space will generally have more than two dimensions} covered by the training and validation sets; this guarantees that the model is interpolating at all times (assuming the training and validation set are representative of the behaviour of the system). Obtaining such a dataset is challenging in my many contexts, particularly for offshore environments where conditions are highly variable and measured data for extreme events is rare. Model performance in extrapolation is, therefore, very important in such cases.

\begin{figure}[ht]
  \centering
      \includegraphics[width=1\textwidth]{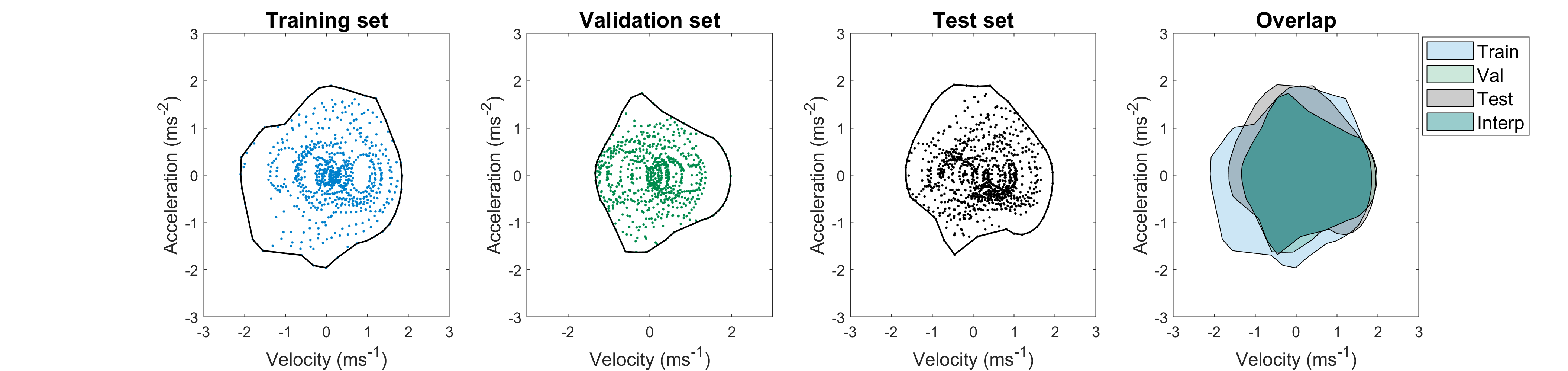}
  \caption{2D visualisations of training, validation and test set boundaries and their overlapping region of interpolation. The black lines represent the boundaries enclosing the coloured points from each dataset. The blue, green and grey shaded regions represent the areas covered by each dataset whilst the shaded teal region represents the area of the test set considered interpolation.}
  \label{fig:Dataset_Overlap}
\end{figure}

For fewer than four dimensions, one may visualise the input space for the training, validation and test sets, and consequently their overlap. Here, considering the input space in terms of the only the velocity and acceleration of the wave particles and not their lags, Figure \ref{fig:Dataset_Overlap} plots the boundaries for the three datasets used in the previous section. The boundaries of the dataset are determined to be the maximum projections from the origin of the input space in all directions encompassed by the measured data. In areas where the testing set overlaps the training and validation sets, the model will be interpolating, else it will be extrapolating. It is worth noting that even the `complete' datasets used here represent only a small proportion of conditions able to be experienced by the structure (as the data were chosen where flow conditions were close to uni-directional, as discussed at the start of Section 3). The 2D boundaries of wave velocity and acceleration were found to be comparable to the boundaries of the first 2 input space PCA components which accounted for 95\% of total variance and deemed representative of the input space as a whole.

\begin{figure}[ht]
  \centering
      \includegraphics[width=1\textwidth]{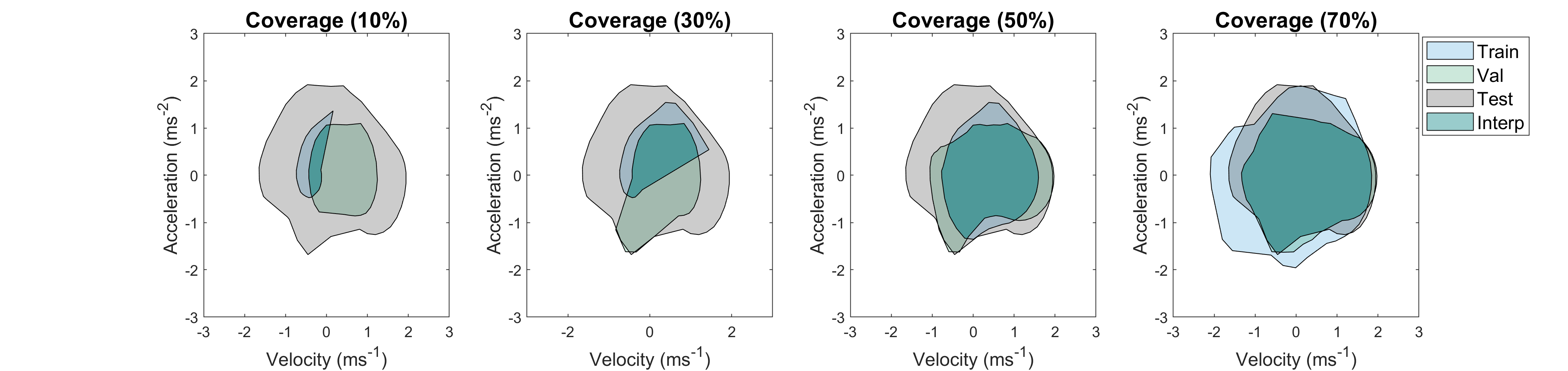}
  \caption{Plots of 10\%, 30\%, 50\% and 70\% test set coverage at increasing quantities of training and validation points. The blue, green and grey shaded regions represent the areas covered by each dataset whilst the shaded teal region represents the area of the test set considered interpolation.}
  \label{fig:Dataset_Coverage_10_30_50_70}
\end{figure}

\begin{figure}[ht!]
  \centering
      \includegraphics[width=0.8\textwidth]{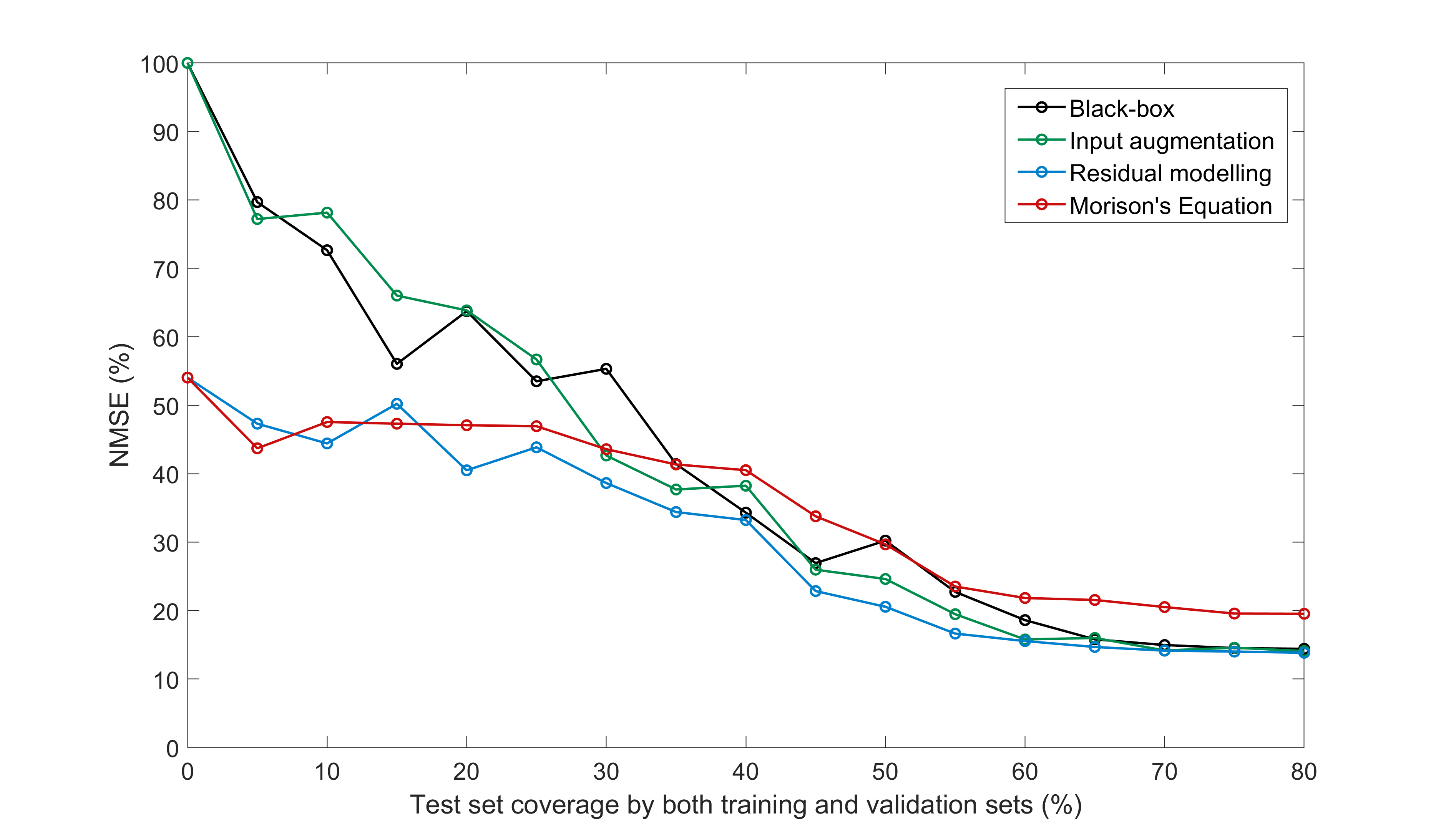}
  \caption{NMSE vs coverage for the MC MPO of black-box and grey-box GP-NARX models and Morison's Equation with all models constructed using the restricted sizes of training and validation sets and prior \(C_d\) and \(C_m\) coefficients suggested by Clauss \cite{Clauss1992Offshore} and DNV-RP-C205 \cite{DNVRPc205}.}
  \label{fig:Coverage_vs_NMSE_Vary_Morison}
\end{figure}

For this two dimensional case, one may use a coverage measure to assess the extent to which one is extrapolating. Here the coverage is calculated as the proportion of area within the test set boundary that lies within the boundaries of both the training and validation sets:
\begin{equation}
\text{Coverage} = 100\left (\frac{A'}{A}\right )
\end{equation}
where \(A'\) is the area within the test set boundary also enclosed by both the training and validation set boundaries and \(A\) is the total area within the test set boundary. Although the density of points within boundaries can vary at each coverage level, a relative measure of extrapolation is achieved, allowing for an investigation in to extrapolative performance of the models. Note that outliers will have a considerable impact on the calculated coverage level, leading to an over estimation of actual coverage and should be removed from datasets where appropriate.
In order to assess model performance at a range of coverage levels, the sizes of the training and validation sets were adjusted to achieve desired levels of coverage. As the number of points used increases, the boundary covered by the datasets will grow and hence cover a larger area of the test set. The rate of growth of the boundary will be highest at low quantities of training and validation points where each additional data point will have a higher chance of widening the boundary. Plots of overlapping boundaries for a range of training and validation set sizes and their respective coverage levels are shown in Figure \ref{fig:Dataset_Coverage_10_30_50_70}. Results of model NMSE from 0\% to 80\% coverage levels in 5\% coverage intervals is shown in Figure \ref{fig:Coverage_vs_NMSE_Vary_Morison}.

The largest difference in model performance is visible at the lowest levels of coverage, with the NMSE of the black-box and input augmentation models increasing steeply as the coverage approaches zero. With no supplied data and zero coverage, all models revert to their prior, resulting in an NMSE of 100\% for both the black-box and input augmentation models. The poor performance of black-models at low levels of coverage is to be expected, however, the input augmentation model failed to offer improvement despite being supplied with the same Morison's Equation prediction and prior \(C_d\) and \(C_m\) coefficients as the residual modelling approach. The physics-based component was unable to assist the input-augmentation model at low coverage due to being supplied as GP input, since the inputs of a GP may only influence a prediction within a proximity to observed data determined by the lengthscale. When far from observed data, the inclusion of Morison's equation as an input will have minimal benefit.
 
The performance of residual modelling suffers significantly less when the coverage is reduced, indicating an improvement in extrapolative capabilities. In the case where no data is supplied, the black-box component of the model reverts to a zero prior, so that the predicted output of the model is now just the prediction of Morison's equation. The usage of Morison's Equation with prior \(C_d\) and \(C_m\) coefficients was able to achieve an NMSE of 54.03\%, a significant improvement over the black-box and input augmentation models.  The white-box acts as a baseline performance for the model which may be improved if data is provided but will not override the improved black-box predictive capabilities in areas where data is available. Residual modelling combined the same white-box and black-box components as the input augmentation model but in a means that achieved superior extrapolative performance.

\begin{figure}[ht!]
  \centering
      \includegraphics[width=0.8\textwidth]{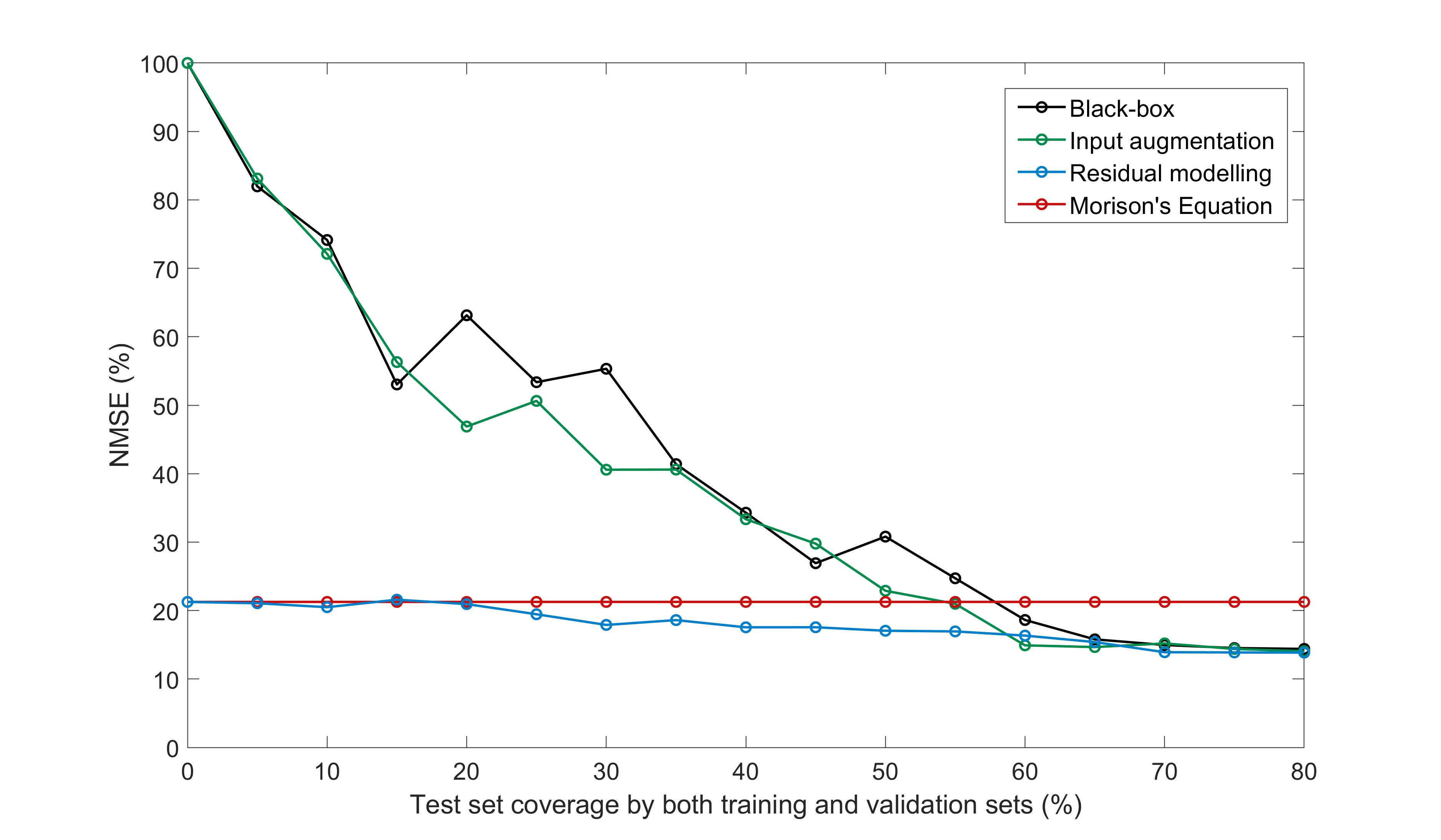}
  \caption{NMSE vs coverage for the MC MPO of black-box and grey-box GP-NARX models and Morison's Equation with \(C_d'\) and \(C_m'\) calculated independently of the supplied training data and kept constant throughout the variation of coverage.}
  \label{fig:Coverage_vs_NMSE_Constant_Morison}
\end{figure}

An alternative investigation in to the effect of coverage on model performance assumed an existing white-box model could be used to assist with predictions, with \(C_d'\) and \(C_m'\) fixed independently of the supplied training data. This scenario  represents the possible case of machine learning implementation within industry where a white-box model is already established and in use. To mimic an established white-box model, the coefficients of Morison's equation were established via Bayesian Linear Regression using the complete validation set, which would not ordinarily be used in white-box model creation, and kept constant throughout the variation of coverage.  Results of model NMSE from 0\% to 80\% coverage levels in 5\% coverage intervals is shown in Figure \ref{fig:Coverage_vs_NMSE_Constant_Morison}.

By incorporating an existing white-box model rather than beginning model construction from scratch, a significant increase in residual modelling performance at low coverage levels can be seen. Where previously, the model performance here was heavily dependant on the selected prior \(C_d\) and \(C_m\) coefficients, the model is now able to revert to the prediction of the existing white-box model, thereby reducing model NMSE to 21.26\% at zero coverage. Provided that the existing model has been validated for use over the intended prediction range, this highlights the benefit of incorporating existing models within the newly created architectures. At high levels of coverage the incorporated existing model has a minimal effect on predictions, with models relying more heavily on their black-box component in areas where data is available. 

\section{Conclusions}
The combining of physics-based white-box and data-based black-box modelling techniques in the form of two grey-box architectures was found to offer benefits in predictive performance over either approach used alone. The best performing grey-box model, the residual modelling GP-NARX, achieved a 29.13\% and 5.48\% relative reduction in NMSE over Morison's Equation and a black-box GP-NARX respectively. It was expected that grey-box models would be of specific help in assisting with extrapolation, an area in which data-based methods typically experience difficulty, and this was indeed found to be the case. Residual modelling was found to offer significant benefits in performance outside the range of observed training conditions, particularly in instances where a pre-established white-box may be available for inclusion in to the combined model.

This paper investigated the combining of Morison's Equation with both GP and GP-NARX regression models to predict wave loading in predominantly unidirectional flow conditions. Further work, incorporating the use of higher fidelity physics models to account for flow in both x and y directions would allow investigation in to prediction capabilities over a wider range of conditions.

\bibliography{Bibliography_wave_Loading}
\newpage
\appendix

\section{Appendix}
Appendix A provides details on the implementation of Bayesian linear regression and determination of the grouped coefficients \(C_d'\) and \(C_m'\) of Morison's Equation along with an overview of Gaussian process regression.
\subsection{Bayesian linear regression}
By gathering the inputs of Morison's Equation, they may be expressed as a single design matrix \(X\) along with model parameters \(\bm{\beta}\): 
\begin{equation}
X = [\bm{U}|\bm{U}|, \ \bm{\dot{U}}]
\end{equation}
\begin{equation}
\bm{\beta} = \begin{bmatrix} C_d'\\ C_m' \end{bmatrix}
\end{equation}
Morison's equation can then be expressed in matrix form:
\begin{equation}
\bm{F} = X\bm{\beta} + \varepsilon \quad\quad\text{where}\quad\quad \varepsilon \sim \mathcal{N}(0,\sigma_n^2\mathbb{I})
\end{equation}
A Bayesian linear regression can be set up for the model.
\begin{equation}
p(\bm{F}|X,\bm{\beta},\sigma_n^2)= \mathcal{N}(X\bm{\beta},\sigma_\beta^2)
\label{eq:BayesLinRegMorison}
\end{equation}
In order to retrieve the desired posteriors over the parameters for Morison's Equation \(\bm{\beta}\) and noise variance \(\sigma_n^2\), it is necessary to place priors over the parameters. Here a Normal-Inverse-Gamma semiconjugate prior is used: 
\begin{equation}
p(\bm{\beta})= \mathcal{N}(m_\beta,\sigma_\beta^2)
\end{equation}
\begin{equation}
p(\sigma_n^2)= \mathcal{IG}(a,b)
\end{equation}
A Gaussian prior over the parameters for Morison's Equation allows for a positive or negative mean \(m_\beta\) with a given variance \(\sigma_\beta^2\). The selection of appropriate \(C_d'\) and \(C_m'\) priors can be made using the dimension specific terms for the structure and standards relating to flow specific drag \(C_d\) and inertia \(C_m\) coefficients such as DNV-RP-C205 \cite{DNVRPc205}. An Inverse Gamma (\(\mathcal{IG}\)) prior can encode belief about the noise variance through hyperparameters \(a\) and \(b\), whilst restricting to only positive values.

The full joint posterior \(p(\bm{\beta},\sigma_n^2|\bm{F},X)\) is unavailable in closed form and it is therefore necessary to calculate the conditional posterior for each parameter: \(p(\bm{\beta}|\bm{F},X,\sigma_n^2)\), \(p(\sigma_n^2|\bm{F},X,\bm{\beta})\).
The parameter posterior distributions were recovered via Gibbs sampling with 10,000 draws. This provided a computationally efficient method for accurate estimation of the conjugate conditional distributions \cite{GibbsSampling}. Once the parameter distributions were retrieved, further sampling was used to recover the posterior distribution of the test set force prediction from (\ref{eq:BayesLinRegMorison}).

\subsection{Gaussian Process Regression}
A Gaussian process (GP) is comprised of a set of random variables, any finite number of which, share a joint Gaussian distribution. A popular view of a GP is as a distribution over functions, where each individual draw from the GP represents a realisation of one of the infinitely many potentially generated functions.

A Gaussian process is completely defined by its mean \(m(x)\) and covariance \(k(x,x')\) functions. The primary component of interest is typically the covariance function with the mean function often taken as zero. For a pair of inputs \(x\) and \(x'\) a GP may be written:
\begin{equation}
f(x) \sim \mathcal{GP}(m(x),k(x,x'))
\end{equation}
Upon observing a set of training points \(X\) of sample size \(N\), a GP is realised as a joint Gaussian distribution of dimensionality \(N\). For a prediction \(\bm{y^*}\) at new set of test points \(X_{\star}\), the covariance matrices for the training points \(K(X,X)\), the test points \(K(X_{\star},X_{\star})\) and between the training and test points \(K(X,X_{\star})\) are required. For predictions based on noisy observations with an assumed Gaussian noise of variance \(\sigma_n^2\), the problem may be formulated:
\begin{equation}
\begin{bmatrix}
\bm{y}\\ \bm{y^*}
\end{bmatrix} \sim \mathcal{N}
\begin{pmatrix}
\bm{0}, & \begin{bmatrix}
K(X,X) + \sigma^2_n \mathbb{I}& K(X,X_{\star})\\ 
K(X_{\star},X) & K(X_{\star},X_{\star}) + \sigma^2_n \mathbb{I}
\end{bmatrix}
\end{pmatrix}
\end{equation}
Expressions for the expected mean and variance of the predicted target \(\bm{y^*}\) may then be derived \cite{GPRasmussen}:
\begin{equation}
\mathbb{E}(\bm{y^*}) = K(X_{\star},X_{\star})(K(X,X)+\sigma^2_n I)^{-1}\bm{y}
\end{equation}
\begin{equation}
\mathbb{V}(\bm{y^*}) = K(X_{\star},X_{\star})-K(X_{\star},X)(K(X,X)+\sigma^2_n I)^{-1}K(X,X_{\star}) + \sigma^2_n \mathbb{I}
\end{equation}
Note here that for the predictive distribution a noisy test target, the noise variance \(\sigma_n^2\) should be accounted for within the expected variance.\\
In this work a squared exponential kernel with Automatic Relevance Determination (ARD) is used. This allows an independent length scale for each input parameter and offers increased model flexibility when operating with multiple types of input parameter.
\begin{equation}
k(x_i,x_j) = \sigma_f^2 \: \text{exp}\left (-\frac{1}{2}(x_i-x_j)^T \Lambda^{-1}(x_i-x_j)\right )
\end{equation}
where \(\sigma_f^2\) is the signal variance and \(\Lambda\) is the matrix of length scales such that \(diag(\Lambda)=[l_1^2,l_2^2,...,l_D^2]\) for a \(D\) dimensional input. These parameters are typically optimised over the Negative Log Marginal Likelihood (NLML) of model predictions on a training set:
\begin{equation}
-log\,p(\bm{y}|X,\bm{\theta})= \frac{1}{2}\bm{y}^T(K+\sigma^2 I)^{-1}\bm{y} + \frac{1}{2}log|K+\sigma^2 I| + \frac{n}{2}log(2\pi)
\end{equation}

As discussed in the main body of the paper, a more appropriate choice of cost function for a GP-NARX is the Negative Log Predictive Likelihood of the Model Predicted Output (MPO NLPL) on an independent validation set. Here, a training set is used to construct a GP-NARX model with the MPO generated by propagating through validation set time steps. The process is defined in Algorithm \ref{Alg:MPONLPL}.

The NLPL of the MPO is calculated as a joint Gaussian likelihood of each measured data point \(y_t\) coming from the corresponding predictive distribution \(y_{t}^*\sim \mathcal{N}(\mathbb{E}(y_{t}^*),\mathbb{V}(y_{t}^*))\) of the GP-NARX output. The MPO NLPL may be derived as:
\begin{equation}        
-log\,p(\bm{y}|\mathbb{E}(\bm{y^*}),\mathbb{V}(\bm{y^*}),\bm{\theta}) = \frac{1}{2}(\bm{y}-\mathbb{E}(\bm{y^*}))^T\mathbb{V}(\bm{y^*})^{-1}(\bm{y}-\mathbb{E}(\bm{y^*})) + \frac{1}{2}\sum_{t=1}^{n}log(\mathbb{V}(y^*_{t})) + \frac{n}{2}log(2\pi)                
\end{equation}

\section{GP-NARX Implementation for CBT dataset}
Appendix B details specifics of GP-NARX implementation including the optimisation of hyperparameters, selection of lagged terms within the model and a comparison of computation time with other models.
\subsection{Hyperparameter optimisation} 
Quantum Behaved Particle Swarm Optimisation (QPSO) \cite{QPSO} was used as a global, gradient-free method for the determination of hyperparameters in GP covariance functions, although it would be possible to use any other appropriate optimisation scheme. To ensure stable convergence, optimisation runs were repeated and the hyperparameters cross checked. The swarm size and cost function stability tolerance were adjusted accordingly until stability was achieved over 12 repeated optimisation runs. The required parameter settings for stable convergence of the GP and GP-NARX models are shown in Table \ref{tab:QPSO}.
\begin{table}[ht]
	\sisetup{round-mode=places,round-precision=3}
	\begin{center}
		\caption{QPSO parameter settings used for GP and GP-NARX optimisation.}
		\label{tab:QPSO}
		\begin{tabular}{lSc} 
			\toprule
			\textbf{Model type} & \textbf{Swarm size (n)} & \textbf{Cost function stability tolerance (t)}\\
			\midrule
			GP & 200 & \(1\times 10^{-3}\)\\
			GP-NARX & 1000 & \(1\times 10^{-5}\)\\
			\bottomrule
		\end{tabular}
	\end{center}
\end{table}

In order to achieve stable convergence of hyperparameters, the GP-NARX required both a higher swarm size and a tighter cost function stability tolerance than the static GP. There were two major reasons for this: the increased number of hyperparameters introduced via additional length scales for lagged inputs and the increased complexity of the cost function. The additional hyperparameters increased the dimensionality of the search space whilst the propagation present within the calculation of the MPO NLPL cost function led to a high sensitivity to changes in hyperparameters.
\subsubsection{GP-NARX lag selection}
The selection of lag terms within a GP-NARX model has a significant effect on the structure and performance of the model. The number of lags included for both the previous signal values and exogenous inputs can be optimised by considering \(l_{u}\) and \(l_{y}\) as hyperparameters. The optimal model may then be chosen via calculation of an appropriate model selection criterion.

Here we use the Akaike Information Criterion (AIC) \cite{Akaike1974} and Bayesian Information Criterion (BIC) \cite{BIC1978} in combination for the selection of lags, where for the AIC, the second order variant (AIC\textsubscript{c}) \cite{AICCSmallSamp} is appropriate due to the low sample size relative to the number of model parameters.

The primary difficulty faced within GP-NARX lag selection is the computational cost of the search. Even when considering only the maximum lags, rather than the full combinatorial problem, a significant number of models are required to be constructed and evaluated. Due to the long training time of the GP-NARX, particularly when using the MPO NLPL cost function and accounting for repeated optimisation starts, the computational cost of covering even moderate search spaces becomes an issue. A proposed solution is to perform the search on a computationally inexpensive AutoRegressive model with eXogeneous inputs (ARX) and carry the lag selections forward for use in the full GP-NARX model. Although still autoregressive in nature and trained using the same datasets as the GP-NARX, it should be noted that the ARX is a linear model and will thus capture a reduced range of behaviours when compared with the GP-NARX. This will likely introduce slight deviation from the optimal lag selections, however the result is expected to provide a sensible lag selection with performance of the final GP-NARX model close to optimal. The pragmatic decision was taken here to make a compromise between computation time and potential improvements in model performance.

\begin{figure}[ht]
	\centering
	\includegraphics[width=0.7\textwidth]{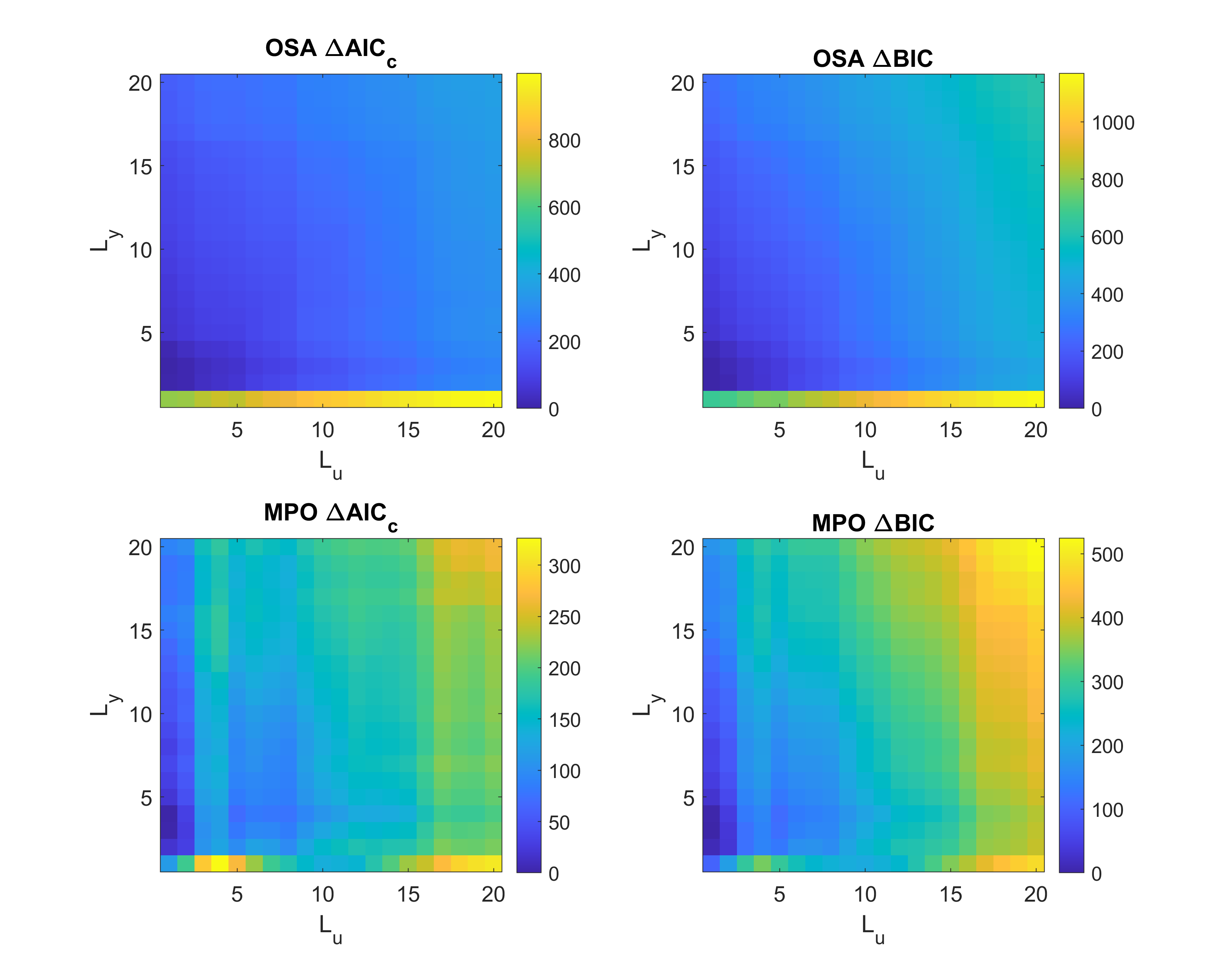}
	\caption{Heatmaps of lag selection metrics for the OSA and MPO predictions of the ARX model.}
	\label{fig:Lag_Heatmap}
\end{figure}

An ARX model is considered for the lag selection search of the form:
\begin{equation}
\bm{y_t} = \sum_{i=0}^{l_u}\alpha _i\bm{u_{t-i}} + \sum_{i=1}^{l_y}\beta _i\bm{y_{t-i}} + \varepsilon
\end{equation}
where similarly to the GP-NARX, the previous signal values, \(\bm{y}\) are the wave force and the exogenous inputs considered are the velocity, \(\bm{U}\) and acceleration, \(\bm{\dot{U}}\) of the wave particles.
\begin{equation}
\bm{u}^T = [\bm{U}|\bm{U}|\bm{_{t}},\ \bm{\dot{U}_{t}},\ \bm{U}|\bm{U}|\bm{_{t-1}},\ \bm{\dot{U}_{t-1}},\ ..., \ \bm{U}|\bm{U}|\bm{_{t-l_u}},\ \bm{\dot{U}_{t-l_u}}]
\end{equation}
\begin{equation}
\bm{\alpha} = [C_{d0}',\ C_{m0}',\ C_{d1}',\ C_{m1}',\ ..., \ C_{dl_u}',\ C_{ml_u}']
\end{equation}
\begin{equation}
\bm{y}^T = [\bm{y_{t-1}},\ \bm{y_{t-2}},\ ..., \ \bm{y_{t-l_y}}]
\end{equation}
\begin{equation}
\bm{\beta} = [C_{y1}',\ C_{y2}',\ ..., \ C_{yl_y}']
\end{equation}
A search space of up to 20 lagged time steps was considered for both the outputs and exogenous inputs. A heatmap of \(\Delta AIC_{c}\) and \(\Delta BIC\) values for both the OSA and MPO prediction of the ARX model is shown in Figure \ref{fig:Lag_Heatmap}. The blue areas represent lower values of \(\Delta_i\) and superior models whilst yellow indicates higher \(\Delta_i\) values and therefore worse models.

The optimum lags were found to be \(l_{u}=1\) and \(l_{y}=3\) for both the OSA and MPO predictions of \(\Delta BIC\) and the OSA prediction of \(\Delta AIC_{c}\). The MPO prediction of \(\Delta AIC_{c}\) narrowly suggested \(l_{u}=1\) and \(l_{y}=4\) as optimal with \(l_{u}=1\) and \(l_{y}=3\) having a \(\Delta AIC_{c}\) of 1.67. This still provided `substantial' \cite{burnham2002} evidence in favour of the lags \(l_{u}=1\) and \(l_{y}=3\) which were therefore selected for the model.

The results suggest that, uniformly, there is little benefit in considering more than one lag for the particle velocity and acceleration terms. Instead, the richer dynamics are expressed through the autoregressive terms for the force.

\subsubsection{Computation time}
The trade-off between performance and computational demand is an important consideration, particularly within industrial applications. Across all models, perhaps unsurprisingly, the general trend is that those that perform best require an increased computation time. For the training and prediction of a single model this ranges from 0.9 seconds for the linear regression, to an average of 81 seconds for GPs and 2 hours 11 minutes for the extreme case of GP-NARX models.\footnote{Runtimes achieved on a laptop with specification: 16GB RAM, Intel i7-9850H processor (6 core, 2.60-4.60GHz)} This difference is only exaggerated when taking in to account repeated optimisation runs to ensure stable convergence of GP and GP-NARX hyperparameters.

There are two major reasons for the considerably higher computation time of the GP-NARX models: the complexity and computational demand of the MPO NLPL cost function and the requirement of MC samples to propagate uncertainty within the output. During optimisation, for any considered approach \cite{NoFreeLunch}, the complexity of the cost function is tied to the computational cost of the search. For the case of the MPO NLPL, the cost function was both slow to evaluate due the requirement of propagation through an independent validation set and highly sensitive to small changes in parameters, thereby creating a complex search space with high numbers of local minima. Hyperparameter sensitivity was a particular issue within the GP-NARX MPO and MC MPO due to the feedback of predictions within the model. This meant that the hyperparameters not only had an effect on predictions at the current time step but again for every instance the prediction was used as a lagged input. The high hyperparameter sensitivity meant that an increased swarm size and tighter convergence tolerance had to be used with QPSO to ensure stable optimisation which further slowed computation time. Although the larger factor within the overall run time of the GP-NARX model, the MPO NLPL cost function only affected the training time of the model, whilst the requirement of MC samples for the propagation of uncertainty affected the prediction time of the model. The computational demand within prediction is a priority for machine learning techniques as models are generally required to be trained once but make predictions repeatedly. In order to reduce prediction time, it would be possible to explore reductions within the number of MC samples used and achieve compromises between predictive stability and computation time.

An alternative means by which to reduce computation cost, with a more specific focus on training time, is through the use of sparse GPs. A subset of training points or set or pseudo-input points \cite{PseudoInputGP} is used to approximate the true posterior of the GP. The computational cost of training is reduced from \(O(n^3)\) to \(O(nm^2)\) \cite{SparseGPTitsias}, where \(n\) is the number of data points and \(m\) is the size of the subset. These methods are most useful in cases of very large datasets, with a training set of \(n=700,000\) and \(m=1000\) being effective for the estimation of flight delay times \cite{GPsForBigData}. Their implementation should be considered if the dataset was expanded to cover a wider range of conditions.
\end{document}